\def\year{2021}\relax
\documentclass[letterpaper]{article} 
\usepackage{aaai22}  
\usepackage{times}  
\usepackage{helvet} 
\usepackage{courier}  
\usepackage[hyphens]{url}  
\usepackage{graphicx} 

\usepackage{color}
\usepackage{amsmath}
\usepackage{amssymb}
\usepackage{microtype}
\usepackage{algorithm, float}
\usepackage{algorithmic}
\usepackage{subfig}
\usepackage{booktabs} 
\usepackage{hyperref}
\usepackage{tablefootnote}
\usepackage{multirow}
\usepackage{subfloat}
\usepackage{makecell}

\usepackage{color}

\newcommand{\vect}[1]{\mathbf{#1}}

\usepackage{natbib}  
\usepackage{caption} 
\title{Learning Wildfire Model from Incomplete State Observations}
\author{
   Alissa Chavalithumrong, Hyung-Jin Yoon, Petros Voulgaris 
}
\affiliations{
     Department of Mechanical Engineering, University of Nevada-Reno\\
     alissachaval@nevada.unr.edu, hyungjiny@unr.edu, pvoulgaris@unr.edu

}

\begin{document}

\maketitle

\begin{abstract} 
As wildfires are expected to become more frequent and severe, improved prediction models are vital to mitigating risk and allocating resources. With remote sensing data, valuable spatiotemporal statistical models can be created and used for resource management practices. In this paper, we create a dynamic model for future wildfire predictions of five locations within the western United States through a deep neural network via historical burned area and climate data. The proposed model has distinct features that address the characteristic need in prediction evaluations, including dynamic online estimation and time-series modeling. Between locations, local fire event triggers are not isolated, and there are confounding factors  when local data is analyzed due to incomplete state observations. When compared to existing approaches that do not account for incomplete state observation within wildfire time-series data, on average, we are able to achieve higher prediction performances.

\end{abstract}
\section{Introduction}

    Wildfire is a vital, dynamic force of nature that shapes the composition and structure of ecosystems. Human interaction with fire has forced it to become both a natural resource management tool, and a hazard that needs to be contained. Anthropogenic-induced climate change, human activity, accumulation of fuels, and past forest management techniques are all contributing factors to the increasing number and severity of forest fires \cite{moritz2014learning}. Despite increased investments in firefighting, the probability and potential losses associated with wildfire risks are increasing and emergency responses are insufficient to protect natural resources and communities \cite{finney2021wildland}. This is reflected on a global scale: North America, South America, Australia, and the Mediterranean Basin have all experienced substantial losses in life and property to wildfires \cite{bowman2011human, chapin2008increasing, holz2017southern, walker2019increasing}. In the western United States alone, the 2020 season saw over 2.5 million hectares burned, the largest wildfire season recorded in modern history \cite{higuera2020record}. Identifying areas with high fire susceptibility is crucial to successfully designing fire management plans and allocating resources \cite{dennison2014large}; consequently, robust approaches to accurately estimate the time, location, and extent of future fires are necessary \cite{sakellariou2019determination, alcasena2019towards}.
    
\begin{figure}[t]
\vskip 0.2in
\begin{center}
\centerline{\includegraphics[width=0.8\columnwidth]{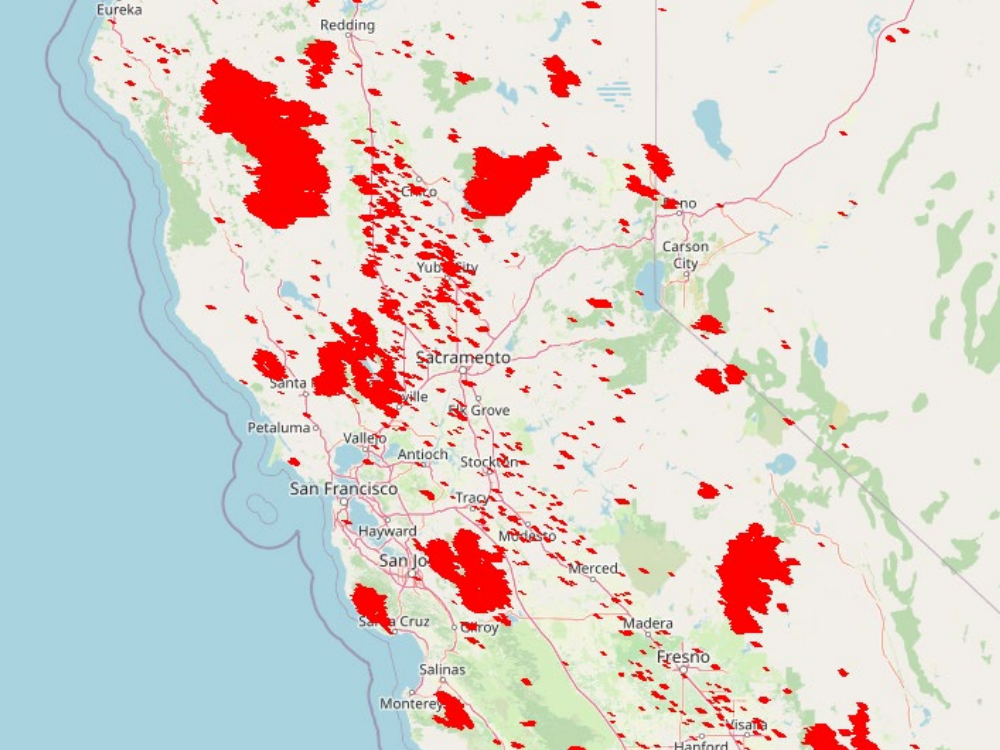}}
\caption{Wildfire in Northern California between 06/01/2020 and 12/31/2020 from Fire Information for Resource Management System (FIRMS) data set~\cite{davies2019nasa} queried from Google Earth Engine~\cite{gorelick2017google}.}
\label{fig/diagram}
\end{center}
\vskip -0.2in
\end{figure}

        Although wildfires have often presented challenges in modeling due to their complex nature, advances in remote sensing techniques, numerical weather prediction, and climate models have allowed for an increase in datacentric approaches \cite{jain2020review}. To make use of these data repositories, spatiotemporal models can be created to extract data and find patterns that may not have been explicitly stored. Early methods included an integrated forecasting framework using a dynamic recurrent neural network, relying on spatiotemporal clustering and classification as the primary data mining technique \cite{cheng2008integrated}. 
  
    Predictive modeling of wildfires often face a limitation that falls in their evaluation phase \cite{gholami2021there}. In supervised machine learning, and specifically classification, the objective is to find a model that accurately assigns data to predefined classes. Commonly, a data set is randomly divided into three parts: a training set, a validation set, and a separate test set to validate the generalized performance of the final classifier \cite{alpaydin2020introduction}. In a wildfire data set, this type of partitioning is not advantageous for future fire prediction. As a result, spatial distribution is valued over temporal aspects, resulting in fire predictions that are not reflective of true burn behavior patterns \cite{jain2020review}. By evaluating our models on both a temporal- and spatial-based division of data in the supervised learning phase, a more realistic future fire prediction model is possible.

\begin{figure}[t]
\begin{center}
\centerline{\includegraphics[width=0.8\columnwidth]{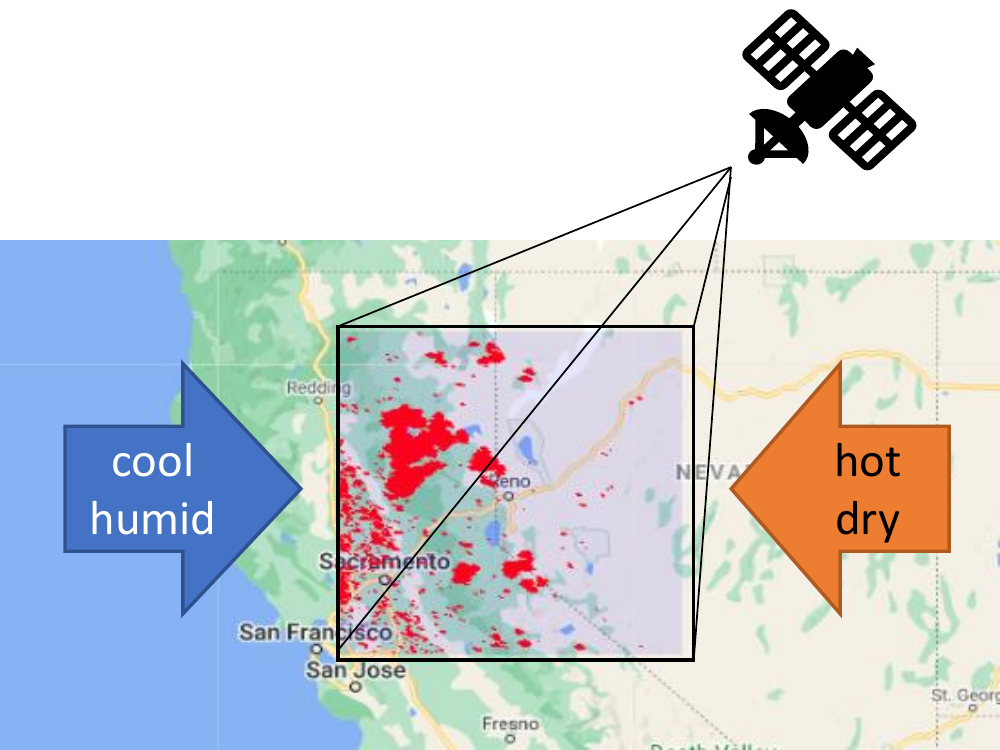}}
\caption{Incomplete state observation of wildfire.  External influence from the surroundings and noisy satellite imagery that observes fire through the atmosphere cause incomplete state observations.}
\label{fig/incomplete_state_observation}
\end{center}
\vskip -0.2in
\end{figure}


In this paper, we work towards creating a wildfire prediction model using remote sensing data by leveraging incomplete state observations, as shown in Figure~\ref{fig/incomplete_state_observation}. In evaluations, we find that our model outperforms baselines against static generative models and recurrent neural network methods, suggesting that satellite imagery contains data with impacted accuracy, and distortions that can result in a partial observation of states. We summarize our contributions as follows:
\begin{itemize}
\item We simulate a real-world data input stream, in which we divide our training data set temporally and create a dynamic model for future wildfire burn area predictions on a weekly temporal resolution.
\item We introduce a hidden Markov model approach that can consider the unknown external effect and hidden state by maximizing the likelihood of the predicted observation, in addition to predicting the fire. 
\end{itemize}

\section{Related Works}

Wildfire forecasting models have been created through several methods, but existing systems, such as the Canadian Forest Fire Weather Index System, rely on weather stations and closed-form mathematical equations to attain a fire risk probability \cite{van1987development}. Recent literature commonly exhibits spatial fire-susceptibility models using either remotely sensed or agency-reported fire data. An example of this is regional fire-susceptibility mapping in Iran with neuro-fuzzy systems \cite{jaafari2019hybrid}. Other notable spatial models include \citet{bashari2016risk} and \citet{shang2020spatially}. Limitations for spatially explicit work are described in the introduction. 

\citet{cheng2008integrated} proposed a spatiotemporal predictive model based on a dynamic recurrent neural network using historical observations in Canada. In a deep learning model proposed by \citet{49935}, remote sensing data of historical fire data are aggregated into a convolutional auto-encoder and LSTM models to estimate fire likelihood. \citet{gholami2021there} uses temporally divided data to test both linear and nonlinear models for fire prediction.

Prior literature have traditionally formulated the fire prediction problem as a classification problem. \citet{shidik2014predicting} used a back-propagation neural network to predict burn areas, achieving an extremely high accuracy compared to other models, which performed with much lower accuracy \cite{coffield2019machine, mitsopoulos2017data}. These lower accuracy works, however, created classes based upon the ground-truth burned area size instead of clusters on the covariates. Similarly, a study done using Australian bush-fires as the primary data set utilized a classifier ensemble method to establish a relationship between fire incidence and climatic data on a weekly time scale \cite{dutta2016big}.

    \begin{figure}[t]
    \vskip -0.25in
    \begin{center}
    \centerline{\includegraphics[width=\columnwidth]{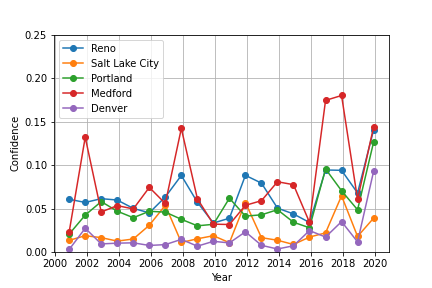}}
    \caption{Wildfire trend observed from the fire detection confidence averaged over 2D pixels and 3 months time span from FIRMS~\cite{davies2019nasa}.}
    \label{fig/diagram}
    \end{center}
    \vskip -0.2in
    \end{figure}

\citet{jin2020ufsp} uses V graph-convolutional layers in a custom architecture to predict urban fires. By integrating a graph convolutional neural network, CNN layers, and RNNs to model the fire prediction problem, the model creates a fully represented and connected graph of the regional data set. Training such a network would require a significant amount of time, and it is likely computationally infeasible to use this model or replicate such results over larger areas.

There are other existing machine-learning-based approaches in the domain of forest fires but they do not specifically focus on fire prediction. \citet{yoon2021estimation} demonstrates a method to determine state estimation within a spatial temporal grid map for UAV wildfire detection. \citet{wijayanto2017classification} used an adaptive neuro fuzzy interface system to predict forest fire occurrences and explore the relationship of fire with human activities. Due to the characteristics of the fire models, ecological prediction models can be analogous as well. \citet{jones2021iteratively} uses an iterative modeling cycle to forecasting biological invasions. \citet{van2020spatially} uses random forest and logistic regression to predict seasonal rainfall induced landslides.


\begin{figure*}[h]
\centering
\subfloat[][Mean air temp.~$(K)$]{\includegraphics[width=0.2\linewidth]{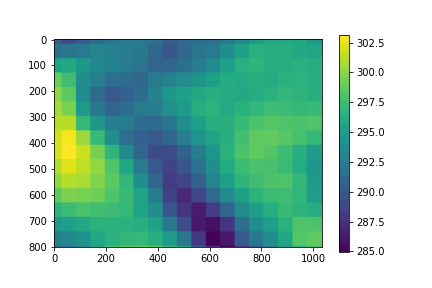}
}
\subfloat[][Dewpoint temp.~$(K)$]{\includegraphics[width=0.2\linewidth]{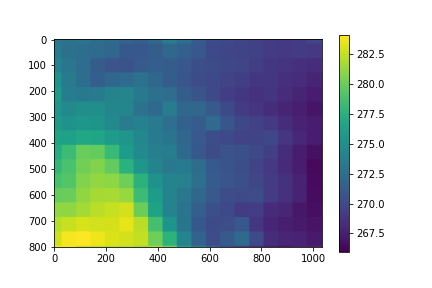}
}
\subfloat[][Total precipitation~$(m)$]{\includegraphics[width=0.2\linewidth]{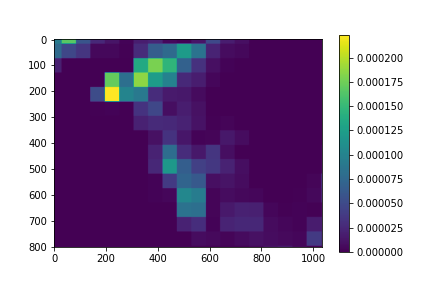}
}
\subfloat[][Surface pressure~$(Pa)$]{\includegraphics[width=0.2\linewidth]{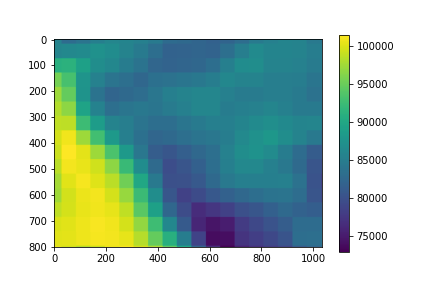}
}
\subfloat[][Wind speed~$(m/s)$]{\includegraphics[width=0.2\linewidth]{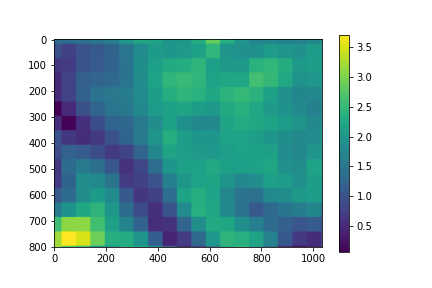}
}
\caption{ERA5 Daily Aggregates~\cite{copernicus2017copernicus}}
\label{fig:era5}
\end{figure*}

\begin{figure}[t]
\begin{center}
\centerline{\includegraphics[width=1.0\columnwidth]{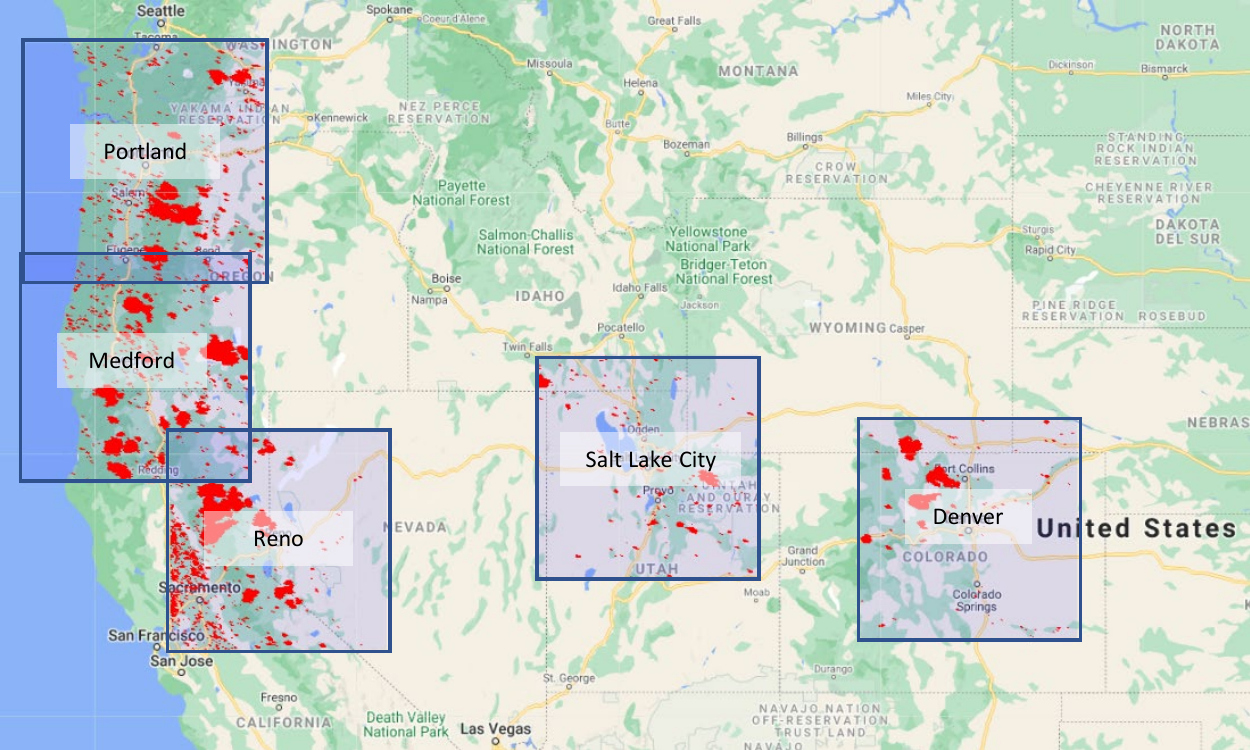}}
\caption{Selected rectangular grids map for data set generation from FIRMS~\cite{davies2019nasa}. The rectangles are centered at Portland, Medford, Reno, Denver,  and Salt Lake City.}
\label{fig/gridmap}
\end{center}
\vskip -0.2in
\end{figure}

\section{Study Area and Data}

In this study, the regions chosen encompass 100 km by 100 km grid squares in the western United States, shown in Figure~\ref{fig/gridmap}. Squares were centered at cities that have experienced moderate to high wildfire activity in the past decade, including: Portland, Oregon; Medford, Oregon; Reno, Nevada; Denver, Colorado; and Salt Lake City, Utah. Wildfire trends in each city are illustrated in Figure~\ref{fig/diagram}. Within the western United States, the Federal government has designated a higher proportion of land for public access, such as national parks and national forests. Consequently, population growth occurring at wildland-urban interfaces has been introduced to these fire hazards \cite{cannon2009increasing}. These cities, due to their increasing population size, proximity to fire zones, and distribution across the American west, provide a diverse and relevant data set to work with. 

\begin{figure}[t]
\begin{center}
\centerline{\includegraphics[width=0.8\columnwidth]{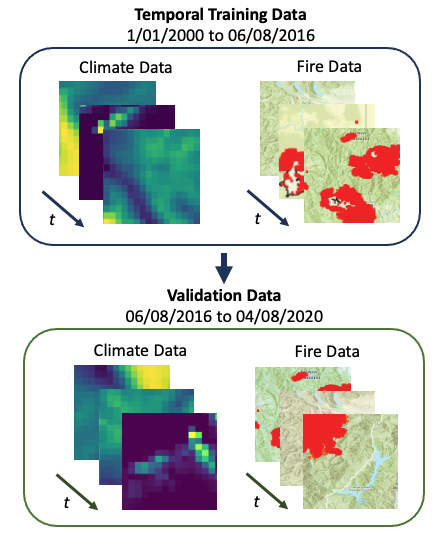}}
\caption{Data partitioning method: data set is divided temporally in weekly time intervals, \textit{t}, following a ratio of 70\% training data and 30\% validation data. This data generation scheme holds true for all locations. }
\label{fig/data}
\end{center}
\vskip -0.2in
\end{figure}

Data was compiled from multiple remote sensing data sources from Google Earth Engine \cite{gorelick2017google}, a publicly available data repository. We selected sources with high spatial and temporal resolution, extensive geographical and historical coverage, and regular update intervals. Wildfire data is from the FIRMS: Fire Information for Resource Management System data set, containing the LANCE fire detection product in rasterized form that provides near real-time fires detection \cite{davies2019nasa}. Climate data is from ERA5 data set produced by Copernicus Climate Change Service, which consists of aggregated values for each day for temperature, dew point, surface pressure, and precipitation, as shown in Figure~\ref{fig:era5} \cite{copernicus2017copernicus}. ERA5 data is available from 1979 to three months from real-time. The combined data was then partitioned into training and validation data sets, demonstrated in Figure~\ref{fig/data}.

\section{Proposed Framework}

\subsection{Modeling Process}

The dynamic auto-encoder consists of two steps: (1) system identification in which the state estimate is determined by maximising the likelihood of its one-step (one week) future prediction; and (2) fire prediction, in which the ground truth is predicted after four time steps (one month). 

    \begin{figure}[h]
    \vskip 0.2in
    \begin{center}
    \centerline{\includegraphics[width=1.0\columnwidth]{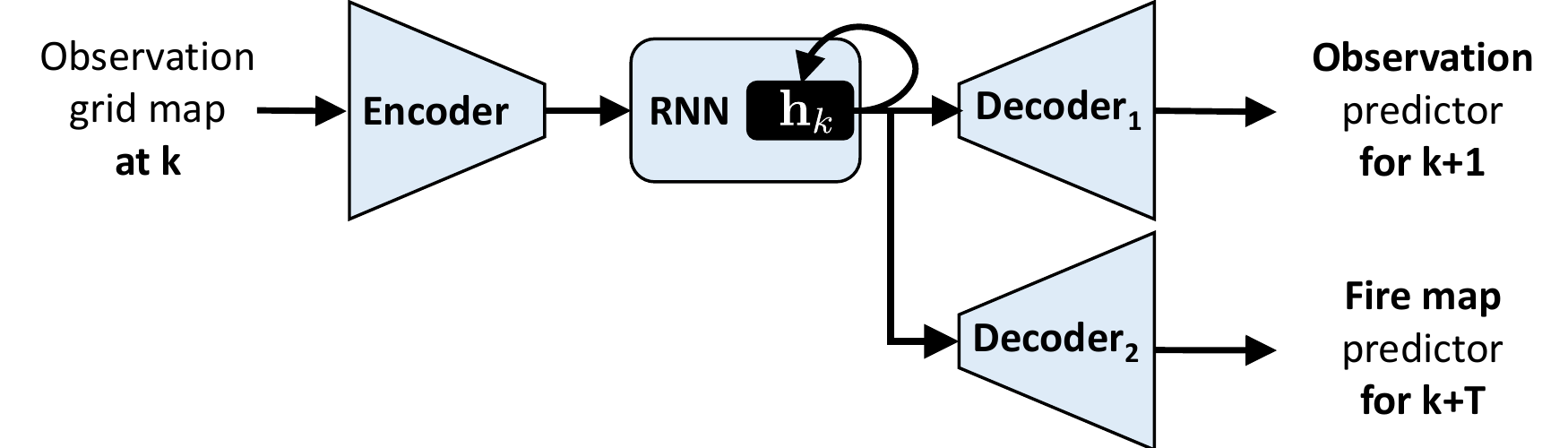}}
    \caption{Dynamic auto-encoder and fire map prediction network. The dynamic auto-encoder is trained to predict the observation at $k+1$ using the observation at $k$ and the state of RNN denoted as $h_k$. And the state of RNN denoted as $h_k$ is also used to predict fire map at $T$ time steps ahead.  }
    \label{fig/diagram_dynamic_autoenc}. 
    \end{center}
    \vskip -0.2in
    \end{figure}

    The dynamic auto-encoder used in this study employs a multidimensional time series from the curated 11-channel data set as the inputs for the model. The encoder transforms the input and feeds the signal to the recurrent neural network (RNN). The hidden state of the RNN was used by the decoder to reconstruct the ground truth input data and estimate the future fire spot likelihood at subsequent time steps. An independent loss function was optimized for the auto-encoder to evaluate the accuracy of the predictions at the given time steps. Further, the RNN state estimate consequently represents the compressed information from the history of available and missing observations and unmeasured variables with high accuracy. Due to the real-time advantage of the recursive update of the dynamic auto-encoder, we can predict the fire grid map online. As shown in Figure~\ref{fig/diagram_dynamic_autoenc}, the dynamic auto-encoder integrates previous observations into the hidden state of RNN. Hence, the model's inference can be implemented in real-time. 
    



        
    
    
    
    
    


\subsection{Online Parameter Estimation}\label{sec:SysID}

The system identification provides the state estimate information for the fire map prediction. As a result, the update of the parameter for system identification is set to be slower than the parameter update for the fire map prediction so that the fire prediction network can track the change of the state estimator of the system identification. In \citet{heusel2017gans}, it was shown that the \emph{Adam}~\cite{kingma2014adam} step size rule can be set to implement the \textbf{two} \textbf{time} scale step size \textbf{update rule} (\emph{TTUR}) in~\eqref{eq:step-size-rule}. 

The two-time scale optimization has update iterations as follows:
\begin{equation}\label{eq:multi-time-scale}
\begin{aligned}
    \theta^\text{pred}_{n+1} &= \theta^\text{pred}_{n} + \epsilon_n^\text{dqn} S_n^\text{pred}(\mathcal{D}_n^\text{trajectory})\\
    \theta^\text{sys}_{n+1} &= \theta^\text{sys}_{n} + \epsilon_n^\text{sys} S_n^\text{sys}(\mathcal{D}_n^\text{trajectory}),
\end{aligned}
\end{equation}
where the step sizes vanish following the vanishing step size rules:
\begin{equation}\label{eq:step-size-rule}
    \epsilon_n^\text{sys}/\epsilon_n^\text{pred} \rightarrow 0 \quad
    \text{as} \quad n \rightarrow \infty,
\end{equation}
and $S_n^\text{sys}$ was defined in~\eqref{eq:sys_id}, and $S_n^\text{pred}$ will be described in~\eqref{eq:pred}. 
\subsubsection{System Identification}
The deep neural networks in Fig.~\ref{fig/diagram_dynamic_autoenc} are trained to predict the future observation in the next time step given previous observations. We use replay buffer~\cite{zhang2017deeper} to save recent trajectories to sample minibatch samples for training as follows:
\begin{equation*}
\mathcal{M}_{\text{trajectory}} =
\begin{bmatrix}
(\mathbf{y}_{0}, \mathbf{h}_{0})_1, &\dots, &(\mathbf{y}_\text{term}, \mathbf{h}_\text{term})_1\\
 \vdots & \vdots & \vdots    \\
(\mathbf{y}_{0}, \mathbf{h}_{0})_{N}, &\dots, &(\mathbf{y}_\text{term}, \mathbf{h}_\text{term})_{N}
\end{bmatrix}
\end{equation*}
where $\mathbf{y}_k$ denotes the observations of the map at time $k$ that are extracted from the climate~\cite{copernicus2017copernicus} and fire data set\cite{davies2019nasa}.   

    \begin{figure}[t]
    \vskip -0.05 in
    \begin{center}
    \centerline{\includegraphics[width=1.0\columnwidth]{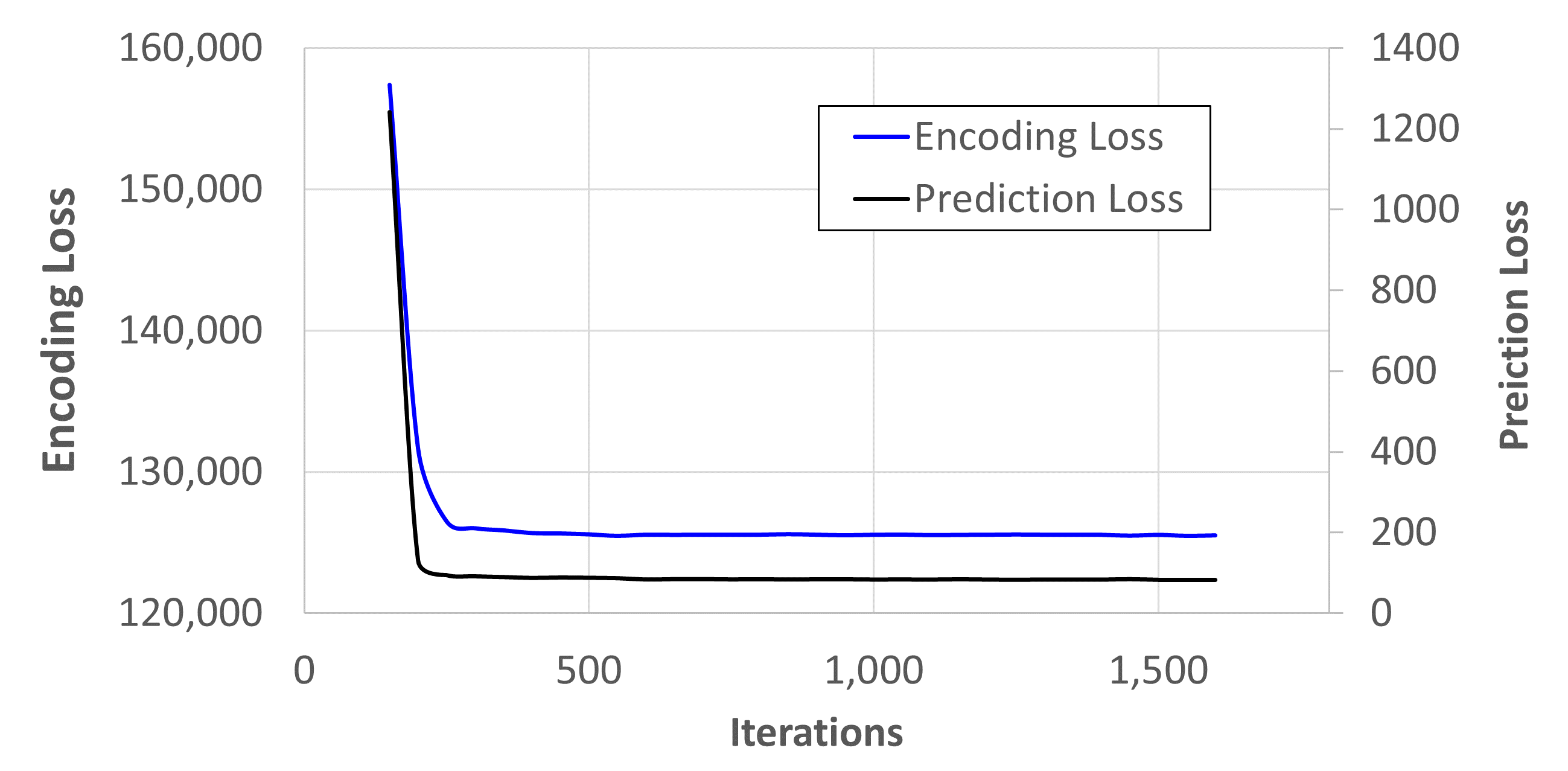}}
    \caption{Training iteration. }
    \label{fig/iteration}
    \end{center}
    \vskip -0.2in
    \end{figure}

The system identification aims to obtain the parameter, which maximizes the likelihood of the one-step future prediction of the observation given the state estimate calculated from the learned model. We maximize the likelihood by minimizing a distance between true image streams and the predicted image streams by a stochastic optimization, which samples trajectories saved in the replay buffer denoted $\mathcal{M}_\text{trajectory}$. The stochastic gradient for the optimization is calculated as follows:    
\begin{equation}\label{eq:sys_id}
    S^\text{sys}_n =  - \nabla_{\theta_\text{sys}} l^\text{sys}(\mathcal{D}_n^\text{trajectory};\theta_\text{sys}),
\end{equation}
where $\mathcal{D}_n^\text{trajectory}:=\{\{\mathbf{y}_k^1, \mathbf{h}_k^1 \}_{k=1}^K, \dots, \{\mathbf{y}_k^L, \mathbf{h}_k^L\}_{k=1}^K\}$ denotes sampled $L$ trajectories from the memory buffer $\mathcal{M}^\text{trajectory}$,  $\{\mathbf{y}_k^l\}_{k=1}^K$ denotes $l$\textsuperscript{th} sample image stream with length $K$, and $\theta_\text{sys}$ denotes the parameter of the model in Fig.~\ref{fig/diagram_dynamic_autoenc}. The loss function $l^\text{sys}(\cdot)$ is calculated as follows:
\begin{equation*}\label{eq:loss_full_observation}
    l^\text{sys}(\mathcal{D}_n^\text{trajectory};\theta_\text{sys}) = \frac{1}{LK}\sum_{l=1}^L\sum_{k=1}^K d(\mathbf{y}_k^l, \hat{\mathbf{y}}_k^l),
\end{equation*}
where $d(\cdot, \cdot)$ denotes a metric~\footnote{In this work, we used the cross-entropy as the distance between predicted and ground truth image.} between two images and $\hat{\mathbf{y}}^l_k$ is the predicted observation grid map (or image frame) for the target $\mathbf{y}^l_k$, and $k$ denotes the time index, and the superscript $l$ denotes the sample index. Each sample trajectory $\{\mathbf{y}_k, \mathbf{h}_0\}$ is processed through the encoder, RNN, and the decoder with the following process. For the image stream sample $\{\mathbf{y}_1, \dots, \mathbf{y}_K\}$ and initial observation $\mathbf{y}_0$ and initial state value $\mathbf{h}_0$, we calculate
\begin{equation}\label{eq:dynautoenc}
    \begin{aligned}
    \vect{h}_{k+1} &= \text{RNN}(\vect{h}_k, \text{Encoder}(\vect{y}_k))\\
    \hat{\vect{y}}_{k+1} & = \text{Decoder}_1(\vect{h}_{k+1})
    \end{aligned}
\end{equation}
and we collect them into $\{\hat{\vect{y}}_1, \dots, \hat{\vect{y}}_K\}$.
\begin{figure*}[t]
\centering
\subfloat[][Reno ]{\includegraphics[width=0.195\linewidth]{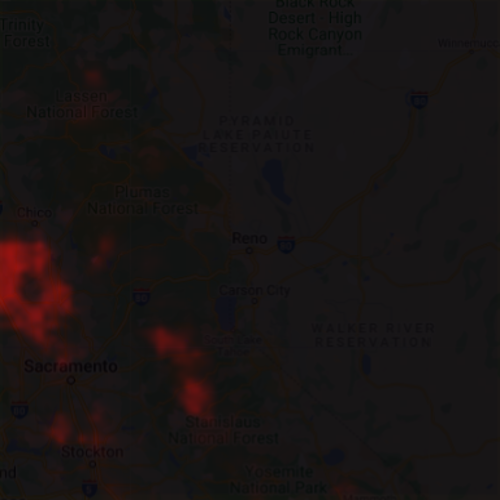}
}
\subfloat[][Salt Lake City]{\includegraphics[width=0.195\linewidth]{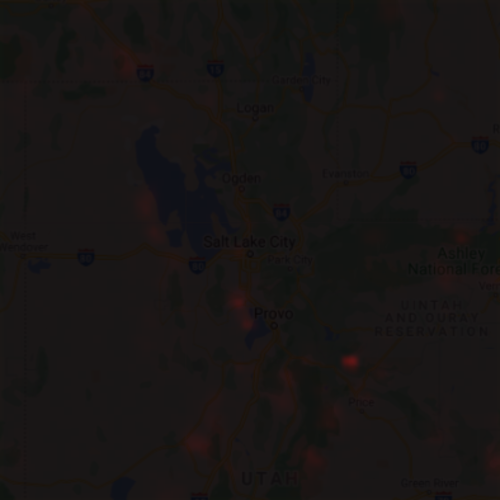}
}
\subfloat[][Portland]{\includegraphics[width=0.195\linewidth]{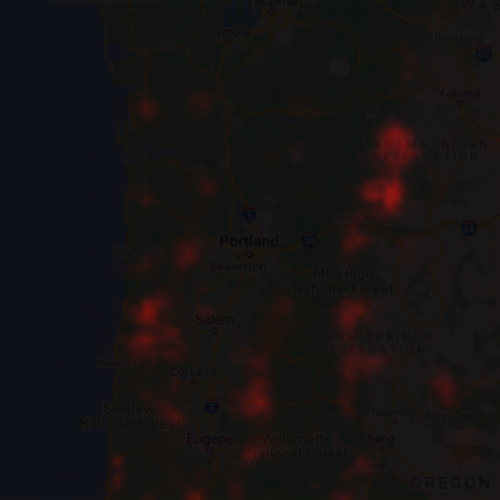}
}
\subfloat[][Medford]{\includegraphics[width=0.195\linewidth]{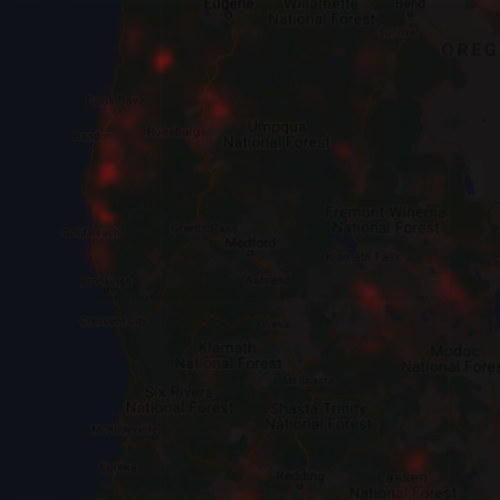}
}
\subfloat[][Denver]{\includegraphics[width=0.195\linewidth]{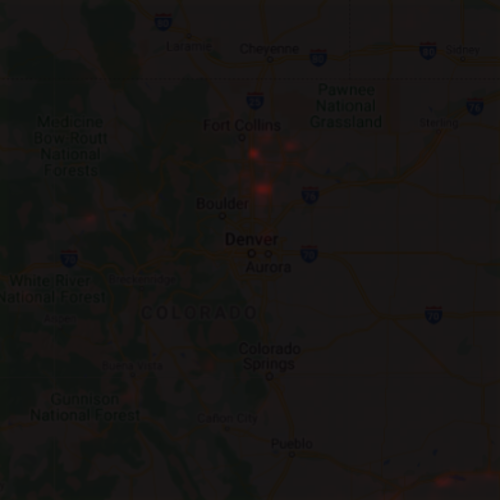}
}
\\
\subfloat[][Reno  (8/23/17)]{\includegraphics[width=0.195\linewidth]{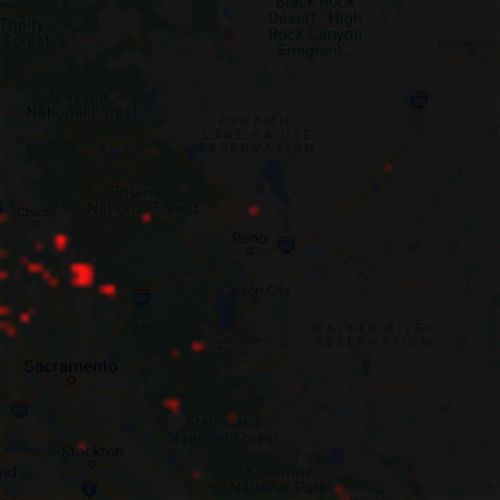}
}
\subfloat[][Salt Lake City (6/20/18)]{\includegraphics[width=0.195\linewidth]{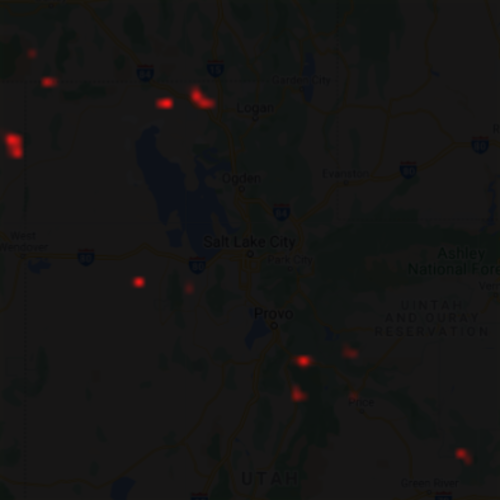}
}
\subfloat[][Portland  (9/19/18)]{\includegraphics[width=0.195\linewidth]{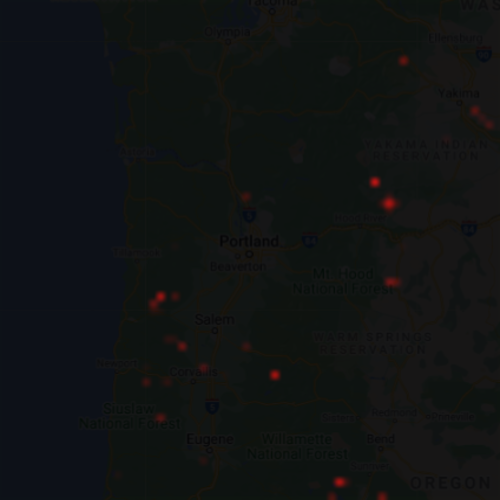}
}
\subfloat[][Medford (7/04/18)]{\includegraphics[width=0.195\linewidth]{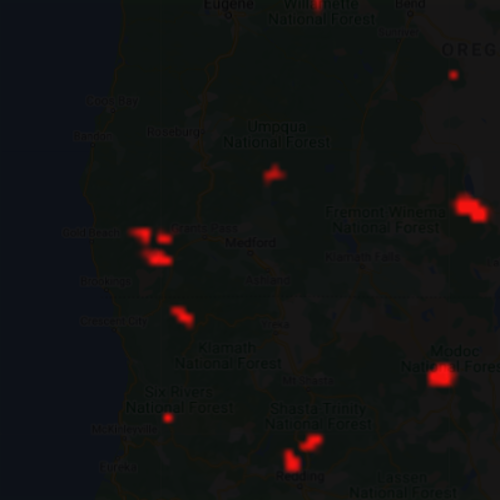}
}
\subfloat[][Denver (12/19/18)]{\includegraphics[width=0.195\linewidth]{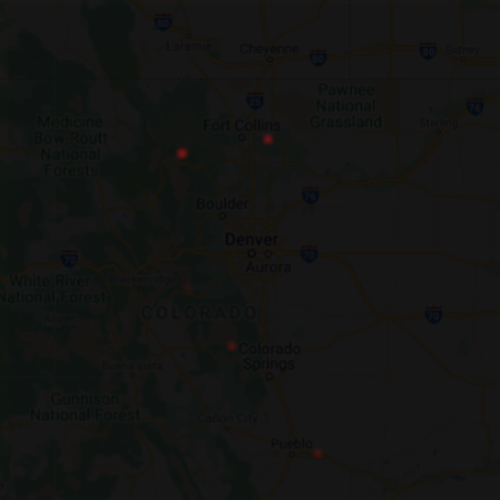}
}
\caption{Fire risk probabilities (top row) predicted 4 weeks before and ground truth fire map~\cite{davies2019nasa} (bottom row).  See the illustrative video linked at \url{https://youtu.be/NjbiGmP3uMk}.}
\label{fig:PredictionMap}
\end{figure*}

\subsubsection{Fire Map Prediction}
The state of the recurrent neural network $\mathbf{h}_k$ is mapped from all previous observations through the recursive update in~\eqref{eq:dynautoenc}. 

\begin{equation}\label{eq:pred}
    S^\text{pred}_n =  - \nabla_{\theta_\text{pred}} l^\text{pred}(\mathcal{D}_n^\text{trajectory};\theta_\text{pred}),
\end{equation}
where the loss function $l^\text{pred}(\cdot)$ is calculated as follows:
\begin{equation*}
    l^\text{pred}(\mathcal{D}_n^\text{trajectory};\theta_\text{pred}) = \frac{1}{LK}\sum_{l=1}^L\sum_{k=1}^K d(\mathbf{f}_{k+T}^l, \hat{\mathbf{f}}_{k+T}^l)
\end{equation*}
with 
\begin{equation*}
    \begin{aligned}
    \vect{h}_{k+1} &= \text{RNN}(\vect{h}_k, \text{Encoder}(\vect{y}_k))\\
    \hat{\mathbf{f}}_{k+T}& = \text{Decoder}_2(\vect{h}_{k}).
    \end{aligned}
\end{equation*}
where $\text{Decoder}_2$ predict $T$ step future fire given the current state estimate $\vect{h}_{k}$.  

We summarise the entire procedure as the following multi-level stochastic optimization in Algorithm~\ref{alg:bi-level-optimization}.

\begin{algorithm}[h]
   \caption{Two-Time-Scale Stochastic Optimization}
   \label{alg:bi-level-optimization}
\begin{algorithmic}
   \STATE {\bfseries Input:} State Estimation Network in Fig.~\ref{fig/diagram_dynamic_autoenc}
   \STATE {\bfseries \hspace{1cm}} with $\theta^\text{sys}_0$ of Encoder, RNN, and Decoder\textsubscript{1}.
   \STATE {\bfseries Input:} Fire map prediction network with  Decoder\textsubscript{2}.
   \STATE {\bfseries Input:} Training time-series data.
   \STATE {\bfseries Input:} Replay buffers: $\mathcal{M}_\text{trajectory}$, $\mathcal{M}_\text{transition}$
   \STATE {\bfseries Output:} Fixed parameters: $\theta^\text{sys}_*$ and $\theta^\text{dqn}_*$
\FOR{$n=0$ {\bfseries to} $N_{\text{iterations}}$}
   \STATE Update RNN state $\mathbf{h}_k$ given $\mathbf{y}_k$ as in ~\eqref{eq:dynautoenc}
   \STATE $\quad \vect{h}_{k+1} \leftarrow \text{RNN}(\vect{h}_k, \text{Encoder}(\vect{y}_k))$
   \STATE $\quad \hat{\vect{y}}_{k} \leftarrow \text{Decoder}(\mathbf{h}_k)$
   \STATE Add new data to the replay buffers
   \STATE $\quad \mathcal{M}_\text{trajectory} \leftarrow (\vect{y}_{k}, h_{k})$
   \STATE Update parameters with the gradients in~\eqref{eq:sys_id} and~\eqref{eq:pred}
   \STATE $\quad \theta^\text{pred}_{n+1} \leftarrow \theta^\text{pred}_{n} + \epsilon_n^\text{pred} S_n^\text{dqn}(\mathcal{D}^\text{trajctory}_n)$
   \STATE $\quad \theta^\text{sys}_{n+1} \leftarrow \theta^\text{sys}_{n} + \epsilon_n^\text{sys} S_n^\text{sys}(\mathcal{D}^\text{trajectory}_n)$
   \STATE Update the step sizes $\epsilon_n^\text{sys}$ and $\epsilon_n^\text{sys}$ using TTUR in~\cite{heusel2017gans}.
\ENDFOR
\STATE \textbf{Fix} parameters with the current ones.
\end{algorithmic}
\end{algorithm}
As shown in Figure~\ref{fig/iteration}, the algorithm simultaneously decreases the two loss function values for the dynamic auto-encoder and fire map predictor.

\begin{figure*}[t]
\centering
\includegraphics[width=\linewidth]{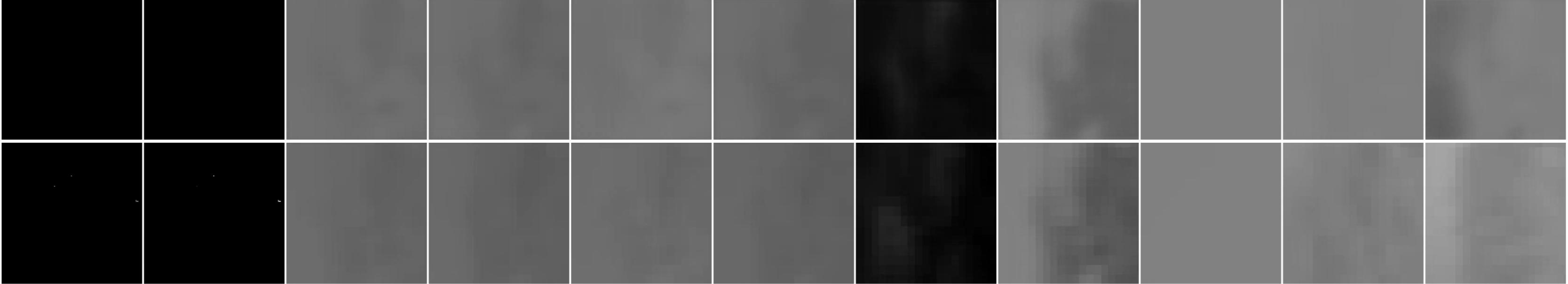}
\caption{One-step future observation prediction for state estimation. The top row shows the prediction and the bottom row shows ground truth. The first two columns are from FIRMS~\cite{davies2019nasa} and the other columns are from ERA5 Daily Aggregates~\cite{copernicus2017copernicus}. All values are scaled to fit within $[0,1]$.}
\label{fig:one-step-prediction}
\end{figure*}

\section{Evaluation and Discussion }

We evaluated the proposed model in Figure~\ref{fig/diagram_dynamic_autoenc} to see if it can generalize over temporal variation. 
We first trained a model using the time-series data from 11/01/2000 to 06/08/2016. The models are trained independently using local time-series data. Then, the corrected trained models were applied to the later time- series validation data beginning 06/08/2016 and ending 04/08/2020.
As shown in Figure~\ref{fig:one-step-prediction} with the linked video, the predicted observation images from the trained model show correlations with ground-truth weather images for the state estimation for the data with incomplete state observations. In Figure~\ref{fig:PredictionMap}, as shown in the linked video, the wildfire prediction map shows qualitative correlation with the ground truth.  The figures and videos demonstrate that the proposed model has prediction capabilities.
\begin{table}[h]
\tiny
\caption{Temporal validation$^2$ for the time-span between 06/08/2016 and 04/08/2020}
\label{tab:validation}
\begin{center}
\begin{small}
\begin{sc}
\begin{tabular}{ccccc}
\toprule
\multirow{2}{*}{Location} & \multirow{2}{*}{Metric} &Gen    &\multirow{2}{*}{GRU} &   Dynamic  \\
                          &                         &Network &                     &  Autoenc  \\
\midrule
\multirow{2}{*}{Reno}  &\textbf{BCE} & \textbf{129.7}  &  134.7  &  140.6     \\
                       &AUROC        &          0.811  &  0.782  &  0.770     \\
\midrule
   Salt                &\textbf{BCE} & $\,\,$90.6   &$\,\,$87.0 &  \textbf{$\,\,$77.1} \\
   Lake City           &AUROC        & 0.663        &  0.734    &  0.765               \\
\midrule
\multirow{2}{*}{Portland} &\textbf{BCE} & 120.8 & 124.9 & \textbf{102.4} \\
                          &AUROC        & 0.743 & 0.803 & 0.676          \\
\midrule
\multirow{2}{*}{Medford} & \textbf{BCE} & 150.74 &  152.0  & \textbf{144.7} \\
                         & AUROC        & 0.698  &  0.645  & 0.734          \\
\midrule
\multirow{2}{*}{Denver} &\textbf{BCE} & 81.3  &\textbf{$\,\,$75.4} & $\,\,$75.8 \\
                        &AUROC        & 0.705 & 0.605              & 0.656      \\

\bottomrule
\end{tabular}
\end{sc}
\end{small}
\end{center}
\footnotesize{$^2$ The training time-series data span from 11/01/2000 and 06/08/2016.}\\
\vskip -0.2in
\end{table}
\subsection{Temporal validation}
We compared the proposed model with two alternatives. The generative network (denoted as GEN. NETWORK) is comprised of an $\text{Encoder}(\cdot)$ and $\text{Decoder}_2(\cdot)$ from Figure~\ref{fig/diagram_dynamic_autoenc}. In the GEN. NETWORK, the $\text{RNN}(\cdot, \cdot)$ is excluded, creating a static model. The gated recurrent unit (denoted as GRU) is comprised of an $\text{Encoder}(\cdot)$, $\text{RNN}(\cdot, \cdot)$ and $\text{Decoder}_2(\cdot)$ but without the $\text{Decoder}_1(\cdot)$, as it will learn with no state estimation.

To validate the proposed model and compare it to alternatives, we computed several performance metrics for processing validation data sets. The validation and training time series data are in different time periods. Therefore, the model did not see the validation data during training. As shown in Table~\ref{tab:validation}, we report standard metrics including area under the receiver operating characteristic (AUROC) and binary cross-entropy loss (denoted as BCE in tables). \emph{The proposed model (dynamic auto-encoder) outperforms the other for the three locations (Salt Lake City, Portland, and Medford) in terms of the binary cross-entropy (BCE) loss.} However, the proposed model did not outperform in terms of the other performance metric. This is partially because AUROC is not a suitable performance metric for the imbalanced dataset ~\cite{weng2008new}. The fire data is imbalanced because grids with active fire on the map are less likely than those without fire. Additionally, the BCE is equivalent to likelihood which is a general statistical model metric. Hence, in our evaluation, we outweigh BCE over AUROC.

The better performance of GRU over GEN NETWORK is expected since the GRU has greater model complexity with its dynamic structure. However, the forward passes from the observation to the prediction of both GRU and the proposed one (DYNAMIC AUTOENC) are identical. This means that the greater performance of the proposed metric could be from the dynamic auto-encoder's ability to estimate for confounding factors.


\section{Conclusion}
In the past decade, wildfires have ravaged the western United States. Due to the scarcity of resources and difficulties in wildfire surveillance, first responders are often a step behind in fire response. In this research, we present a novel, data-driven approach for extracting relevant fire determinants through readily available satellite images. We created an approach to wildfire prediction that explicitly takes state uncertainty into account. We also created a comprehensive model on wildfires that combines both historical fire data with relevant covarying data. Although there are some limitations to our current work, our forecasting model shows optimistic results, and addresses the limitations in current wildfire prediction. As demonstrated by other ecological forecasting topics, to most effectively implement future iterations, resource
managers should be included in collaboration so that through sustained participation further iterations can optimize for real-world decision-making.


\section{Acknowledgments}
We thank the anonymous reviewers for their time and insight. 

\bibliography{mybib.bib}

\begin{thebibliography}{36}
\providecommand{\natexlab}[1]{#1}

\bibitem[{Alcasena et~al.(2019)Alcasena, Ager, Bailey, Pineda, and
  Vega-Garc{\'\i}a}]{alcasena2019towards}
Alcasena, F.~J.; Ager, A.~A.; Bailey, J.~D.; Pineda, N.; and Vega-Garc{\'\i}a,
  C. 2019.
\newblock Towards a comprehensive wildfire management strategy for
  Mediterranean areas: Framework development and implementation in Catalonia,
  Spain.
\newblock \emph{Journal of environmental management}, 231: 303--320.

\bibitem[{Alpaydin(2020)}]{alpaydin2020introduction}
Alpaydin, E. 2020.
\newblock \emph{Introduction to machine learning}.
\newblock MIT press.

\bibitem[{Bashari et~al.(2016)Bashari, Naghipour, Khajeddin, Sangoony, and
  Tahmasebi}]{bashari2016risk}
Bashari, H.; Naghipour, A.~A.; Khajeddin, S.~J.; Sangoony, H.; and Tahmasebi,
  P. 2016.
\newblock Risk of fire occurrence in arid and semi-arid ecosystems of Iran: an
  investigation using Bayesian belief networks.
\newblock \emph{Environmental monitoring and assessment}, 188(9): 1--15.

\bibitem[{Bowman et~al.(2011)Bowman, Balch, Artaxo, Bond, Cochrane,
  D’antonio, DeFries, Johnston, Keeley, Krawchuk et~al.}]{bowman2011human}
Bowman, D.~M.; Balch, J.; Artaxo, P.; Bond, W.~J.; Cochrane, M.~A.;
  D’antonio, C.~M.; DeFries, R.; Johnston, F.~H.; Keeley, J.~E.; Krawchuk,
  M.~A.; et~al. 2011.
\newblock The human dimension of fire regimes on Earth.
\newblock \emph{Journal of biogeography}, 38(12): 2223--2236.

\bibitem[{Cannon and DeGraff(2009)}]{cannon2009increasing}
Cannon, S.~H.; and DeGraff, J. 2009.
\newblock The increasing wildfire and post-fire debris-flow threat in western
  USA, and implications for consequences of climate change.
\newblock In \emph{Landslides--disaster risk reduction}, 177--190. Springer.

\bibitem[{Chapin et~al.(2008)Chapin, Trainor, Huntington, Lovecraft, Zavaleta,
  Natcher, McGuire, Nelson, Ray, Calef et~al.}]{chapin2008increasing}
Chapin, F.~S.; Trainor, S.~F.; Huntington, O.; Lovecraft, A.~L.; Zavaleta, E.;
  Natcher, D.~C.; McGuire, A.~D.; Nelson, J.~L.; Ray, L.; Calef, M.; et~al.
  2008.
\newblock Increasing wildfire in Alaska's boreal forest: Pathways to potential
  solutions of a wicked problem.
\newblock \emph{BioScience}, 58(6): 531--540.

\bibitem[{Cheng and Wang(2008)}]{cheng2008integrated}
Cheng, T.; and Wang, J. 2008.
\newblock Integrated spatio-temporal data mining for forest fire prediction.
\newblock \emph{Transactions in GIS}, 12(5): 591--611.

\bibitem[{Coffield et~al.(2019)Coffield, Graff, Chen, Smyth, Foufoula-Georgiou,
  and Randerson}]{coffield2019machine}
Coffield, S.~R.; Graff, C.~A.; Chen, Y.; Smyth, P.; Foufoula-Georgiou, E.; and
  Randerson, J.~T. 2019.
\newblock Machine learning to predict final fire size at the time of ignition.
\newblock \emph{International journal of wildland fire}, 28(11): 861--873.

\bibitem[{Davies et~al.(2019)Davies, Ederer, Olsina, Wong, Cechini, and
  Boller}]{davies2019nasa}
Davies, D.; Ederer, G.; Olsina, O.; Wong, M.; Cechini, M.; and Boller, R. 2019.
\newblock NASA's Fire Information for Resource Management System (FIRMS): Near
  Real-Time Global Fire Monitoring Using Data from MODIS and VIIRS.
\newblock In \emph{EARSel Forest Fires SIG Workshop}.

\bibitem[{Dennison et~al.(2014)Dennison, Brewer, Arnold, and
  Moritz}]{dennison2014large}
Dennison, P.~E.; Brewer, S.~C.; Arnold, J.~D.; and Moritz, M.~A. 2014.
\newblock Large wildfire trends in the western United States, 1984--2011.
\newblock \emph{Geophysical Research Letters}, 41(8): 2928--2933.

\bibitem[{Dutta, Das, and Aryal(2016)}]{dutta2016big}
Dutta, R.; Das, A.; and Aryal, J. 2016.
\newblock Big data integration shows Australian bush-fire frequency is
  increasing significantly.
\newblock \emph{Royal Society open science}, 3(2): 150241.

\bibitem[{Finney(2021)}]{finney2021wildland}
Finney, M.~A. 2021.
\newblock The wildland fire system and challenges for engineering.
\newblock \emph{Fire Safety Journal}, 120: 103085.

\bibitem[{Gholami et~al.(2021)Gholami, Kodandapani, Wang, and
  Ferres}]{gholami2021there}
Gholami, S.; Kodandapani, N.; Wang, J.; and Ferres, J.~L. 2021.
\newblock Where there’s Smoke, there’s Fire: Wildfire Risk Predictive
  Modeling via Historical Climate Data.
\newblock In \emph{Annual Conference on Innovative Applications of Artificial
  Intelligence (IAAI)}.

\bibitem[{Gorelick et~al.(2017)Gorelick, Hancher, Dixon, Ilyushchenko, Thau,
  and Moore}]{gorelick2017google}
Gorelick, N.; Hancher, M.; Dixon, M.; Ilyushchenko, S.; Thau, D.; and Moore, R.
  2017.
\newblock Google Earth Engine: Planetary-scale geospatial analysis for
  everyone.
\newblock \emph{Remote Sensing of Environment}.

\bibitem[{Heusel et~al.(2017)Heusel, Ramsauer, Unterthiner, Nessler, and
  Hochreiter}]{heusel2017gans}
Heusel, M.; Ramsauer, H.; Unterthiner, T.; Nessler, B.; and Hochreiter, S.
  2017.
\newblock {GAN}s trained by a two time-scale update rule converge to a local
  {Nash} equilibrium.
\newblock In \emph{Advances in neural information processing systems},
  6626--6637.

\bibitem[{Higuera and Abatzoglou(2020)}]{higuera2020record}
Higuera, P.~E.; and Abatzoglou, J.~T. 2020.
\newblock Record-setting climate enabled the extraordinary 2020 fire season in
  the western United States.
\newblock \emph{Global change biology}.

\bibitem[{Holz et~al.(2017)Holz, Paritsis, Mundo, Veblen, Kitzberger,
  Williamson, Ar{\'a}oz, Bustos-Schindler, Gonz{\'a}lez, Grau
  et~al.}]{holz2017southern}
Holz, A.; Paritsis, J.; Mundo, I.~A.; Veblen, T.~T.; Kitzberger, T.;
  Williamson, G.~J.; Ar{\'a}oz, E.; Bustos-Schindler, C.; Gonz{\'a}lez, M.~E.;
  Grau, H.~R.; et~al. 2017.
\newblock Southern Annular Mode drives multicentury wildfire activity in
  southern South America.
\newblock \emph{Proceedings of the National Academy of Sciences}, 114(36):
  9552--9557.

\bibitem[{Huot et~al.(2020)Huot, Hu, Ihme, Wang, Burge, Lu, Hickey, fan Chen,
  and Anderson}]{49935}
Huot, F.; Hu, R.~L.; Ihme, M.; Wang, Q.; Burge, J.; Lu, T.; Hickey, J.~J.; fan
  Chen, Y.; and Anderson, J.~R. 2020.
\newblock Deep Learning Models for Predicting Wildfires from Historical
  Remote-Sensing Data.

\bibitem[{Jaafari et~al.(2019)Jaafari, Zenner, Panahi, and
  Shahabi}]{jaafari2019hybrid}
Jaafari, A.; Zenner, E.~K.; Panahi, M.; and Shahabi, H. 2019.
\newblock Hybrid artificial intelligence models based on a neuro-fuzzy system
  and metaheuristic optimization algorithms for spatial prediction of wildfire
  probability.
\newblock \emph{Agricultural and forest meteorology}, 266: 198--207.

\bibitem[{Jain et~al.(2020)Jain, Coogan, Subramanian, Crowley, Taylor, and
  Flannigan}]{jain2020review}
Jain, P.; Coogan, S.~C.; Subramanian, S.~G.; Crowley, M.; Taylor, S.; and
  Flannigan, M.~D. 2020.
\newblock A review of machine learning applications in wildfire science and
  management.
\newblock \emph{Environmental Reviews}, 28(4): 478--505.

\bibitem[{Jin et~al.(2020)Jin, Zhu, Chen, Sha, Hu, and Huang}]{jin2020ufsp}
Jin, G.; Zhu, C.; Chen, X.; Sha, H.; Hu, X.; and Huang, J. 2020.
\newblock UFSP-Net: a neural network with spatio-temporal information fusion
  for urban fire situation prediction.
\newblock In \emph{IOP Conference Series: Materials Science and Engineering},
  volume 853, 012050. IOP Publishing.

\bibitem[{Jones et~al.(2021)Jones, Jones, Petrasova, Petras, Gaydos, Skrip,
  Takeuchi, Bigsby, and Meentemeyer}]{jones2021iteratively}
Jones, C.~M.; Jones, S.; Petrasova, A.; Petras, V.; Gaydos, D.; Skrip, M.~M.;
  Takeuchi, Y.; Bigsby, K.; and Meentemeyer, R.~K. 2021.
\newblock Iteratively forecasting biological invasions with PoPS and a little
  help from our friends.
\newblock \emph{Frontiers in Ecology and the Environment}.

\bibitem[{Kingma and Ba(2015)}]{kingma2014adam}
Kingma, D.~P.; and Ba, J. 2015.
\newblock Adam: {A} Method for Stochastic Optimization.
\newblock In \emph{International Conference on Learning Representations
  ({ICLR})}.

\bibitem[{Mitsopoulos and Mallinis(2017)}]{mitsopoulos2017data}
Mitsopoulos, I.; and Mallinis, G. 2017.
\newblock A data-driven approach to assess large fire size generation in
  Greece.
\newblock \emph{Natural Hazards}, 88(3): 1591--1607.

\bibitem[{Moritz et~al.(2014)Moritz, Batllori, Bradstock, Gill, Handmer,
  Hessburg, Leonard, McCaffrey, Odion, Schoennagel et~al.}]{moritz2014learning}
Moritz, M.~A.; Batllori, E.; Bradstock, R.~A.; Gill, A.~M.; Handmer, J.;
  Hessburg, P.~F.; Leonard, J.; McCaffrey, S.; Odion, D.~C.; Schoennagel, T.;
  et~al. 2014.
\newblock Learning to coexist with wildfire.
\newblock \emph{Nature}, 515(7525): 58--66.

\bibitem[{Sakellariou et~al.(2019)Sakellariou, Tampekis, Samara, Flannigan,
  Jaeger, Christopoulou, and Sfougaris}]{sakellariou2019determination}
Sakellariou, S.; Tampekis, S.; Samara, F.; Flannigan, M.; Jaeger, D.;
  Christopoulou, O.; and Sfougaris, A. 2019.
\newblock Determination of fire risk to assist fire management for insular
  areas: The case of a small Greek island.
\newblock \emph{Journal of Forestry Research}, 30(2): 589--601.

\bibitem[{Shang et~al.(2020)Shang, Wulder, Coops, White, and
  Hermosilla}]{shang2020spatially}
Shang, C.; Wulder, M.~A.; Coops, N.~C.; White, J.~C.; and Hermosilla, T. 2020.
\newblock Spatially-Explicit Prediction of Wildfire Burn Probability Using
  Remotely-Sensed and Ancillary Data.
\newblock \emph{Canadian Journal of Remote Sensing}, 46(3): 313--329.

\bibitem[{Shidik and Mustofa(2014)}]{shidik2014predicting}
Shidik, G.~F.; and Mustofa, K. 2014.
\newblock Predicting size of forest fire using hybrid model.
\newblock In \emph{Information and Communication Technology-EurAsia
  Conference}, 316--327. Springer.

\bibitem[{Store(2017)}]{copernicus2017copernicus}
Store, C. C. C. S. C.~D. 2017.
\newblock Copernicus Climate Change Service (C3 S)(2017): ERA5: Fifth
  generation of ECMWF atmospheric reanalyses of the global climate.

\bibitem[{Van~Dao et~al.(2020)Van~Dao, Jaafari, Bayat, Mafi-Gholami, Qi,
  Moayedi, Van~Phong, Ly, Le, Trinh et~al.}]{van2020spatially}
Van~Dao, D.; Jaafari, A.; Bayat, M.; Mafi-Gholami, D.; Qi, C.; Moayedi, H.;
  Van~Phong, T.; Ly, H.-B.; Le, T.-T.; Trinh, P.~T.; et~al. 2020.
\newblock A spatially explicit deep learning neural network model for the
  prediction of landslide susceptibility.
\newblock \emph{Catena}, 188: 104451.

\bibitem[{Wagner(1987)}]{van1987development}
Wagner, C. E.~V. 1987.
\newblock Development and structure of the canadian forest fireweather index
  system.
\newblock In \emph{Can. For. Serv., Forestry Tech. Rep}. Citeseer.

\bibitem[{Walker et~al.(2019)Walker, Baltzer, Cumming, Day, Ebert, Goetz,
  Johnstone, Potter, Rogers, Schuur et~al.}]{walker2019increasing}
Walker, X.~J.; Baltzer, J.~L.; Cumming, S.~G.; Day, N.~J.; Ebert, C.; Goetz,
  S.; Johnstone, J.~F.; Potter, S.; Rogers, B.~M.; Schuur, E.~A.; et~al. 2019.
\newblock Increasing wildfires threaten historic carbon sink of boreal forest
  soils.
\newblock \emph{Nature}, 572(7770): 520--523.

\bibitem[{Weng and Poon(2008)}]{weng2008new}
Weng, C.~G.; and Poon, J. 2008.
\newblock A new evaluation measure for imbalanced datasets.
\newblock In \emph{Proceedings of the 7th Australasian Data Mining
  Conference-Volume 87}, 27--32.

\bibitem[{Wijayanto et~al.(2017)Wijayanto, Sani, Kartika, and
  Herdiyeni}]{wijayanto2017classification}
Wijayanto, A.~K.; Sani, O.; Kartika, N.~D.; and Herdiyeni, Y. 2017.
\newblock Classification model for forest fire hotspot occurrences prediction
  using ANFIS algorithm.
\newblock In \emph{IOP Conference Series: Earth and Environmental Science},
  volume~54, 012059. IOP Publishing.

\bibitem[{Yoon et~al.(2021)Yoon, Kim, Shrestha, Hovakimyan, and
  Voulgaris}]{yoon2021estimation}
Yoon, H.-J.; Kim, H.; Shrestha, K.; Hovakimyan, N.; and Voulgaris, P. 2021.
\newblock Estimation and Planning of Exploration Over Grid Map Using A
  Spatiotemporal Model with Incomplete State Observations.
\newblock \emph{arXiv preprint arXiv:2103.02840}.

\bibitem[{Zhang and Sutton(2017)}]{zhang2017deeper}
Zhang, S.; and Sutton, R.~S. 2017.
\newblock A deeper look at experience replay.
\newblock In \emph{Deep Reinforcement Learning Symposium, NIPS 2017}.

\end{thebibliography}

\end{document}